\documentclass{article}
\usepackage{ijcai23}

\usepackage{times}
\usepackage{soul}
\usepackage{xcolor}
\usepackage{url}
\usepackage[hidelinks]{hyperref}
\usepackage[utf8]{inputenc}
\usepackage[small]{caption}
\usepackage{graphicx}
\usepackage{amsmath}
\usepackage{amsthm}
\usepackage{amsmath}
\usepackage{booktabs}
\usepackage{algorithm}
\usepackage{algorithmic}
\usepackage[switch]{lineno}
\usepackage{multirow}
\usepackage{pifont}
\usepackage{multirow}
\usepackage{amsfonts,amssymb}

\title{Towards Robust Scene Text Image Super-resolution\\
via Explicit Location Enhancement}

\author{
Hang Guo$^1$
\and
Tao Dai$^{2,}$\thanks{Corresponding author: Tao Dai}
\and
Guanghao Meng$^{1,3}$ 
\and
Shu-Tao Xia$^{1,3}$
\affiliations
$^1$Tsinghua Shenzhen International Graduate School, Tsinghua University
\\
$^2$College of Computer Science and Software Engineering, Shenzhen University\\
$^3$Peng Cheng Laboratory, Shenzhen, China\\
\emails
\{cshguo, daitao.edu\}@gmail.com,
mgh19@mails.tsinghua.edu.cn,
xiast@sz.tsinghua.edu.cn
}

\begin{document}

\maketitle

\let\thefootnote\relax\footnotetext{This work is supported in part by the National Key Research and Development Program of China under Grant 2022YFF1202104, the National Natural Science Foundation of China under Grant 62171248, Shenzhen Science and Technology Program (Grant No. JCYJ20220818101014030, JCYJ20220818101012025), the PCNL KEY project (PCL2021A07), and Research Center for Computer Network (Shenzhen) Ministry of Education.}


\begin{abstract}
Scene text image super-resolution (STISR), aiming to improve image quality while boosting downstream scene text recognition accuracy, has recently achieved great success. However, most existing methods treat the foreground (character regions) and background (non-character regions) equally in the forward process, and neglect the disturbance from the complex background, thus limiting the performance. To address these issues, in this paper, we propose a novel method LEMMA that explicitly models character regions to produce high-level text-specific guidance for super-resolution. To model the location of characters effectively, we propose the location enhancement module to extract character region features based on the attention map sequence. Besides, we propose the multi-modal alignment module to perform bidirectional visual-semantic alignment to generate high-quality prior guidance, which is then incorporated into the super-resolution branch in an adaptive manner using the proposed adaptive fusion module. Experiments on TextZoom and four scene text recognition benchmarks demonstrate the superiority of our method over other state-of-the-art methods. Code is available at \href{https://github.com/csguoh/LEMMA}{https://github.com/csguoh/LEMMA}.

\end{abstract}

\section{Introduction}
Scene text recognition (STR) is an important computer vision task and has a wide range of applications \cite{liem2018fvi,khare2019novel}. Despite impressive progress made, current STR methods are still struggling with low-resolution (LR) images \cite{WenjiaWang2020SceneTI}. Several approaches \cite{dong1506boosting,tran2019deep} process LR input by treating text images as natural images and employing a generic super-resolution network to obtain high-resolution ones. However, as shown in previous studies \cite{JingyeChen2021SceneTT,chen2022text}, this scheme is not that satisfactory. Therefore, customizing super-resolution networks for scene text images has become a popular research topic.

To improve the quality of LR images, many scene text image super-resolution (STISR) approaches have recently been proposed with promising results. For example, a location-aware loss function is proposed in \cite{JingyeChen2021SceneTT} to consider character spatial distribution. By applying the character probability distribution, TPGSR \cite{JianqiMa2021TextPG} demonstrates the importance of using language knowledge as guidance in the STISR task. To handle spatially irregular text, TATT is proposed in \cite{ma2022text}. Moreover, C3-STISR \cite{MinyiZhao2022C3STISRST} achieves favorable performance by using three perspectives of clues.

\begin{figure}[t]
\centering
\includegraphics[width=0.44\textwidth]{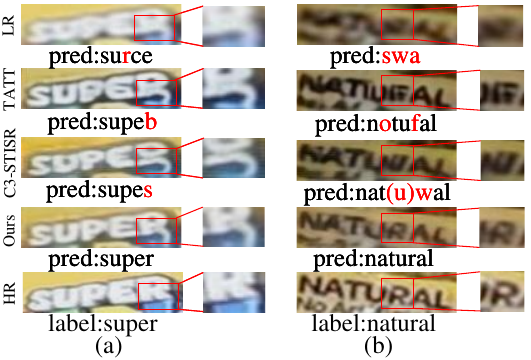}
\caption{Complex background brings challenges to STISR. (a) ``R" in ``SUPER" can be mistakenly reconstructed as ``B" or ``S". (b) Inaccurate character localization due to complex background leads to poor reconstruction.
}
\label{fig1}
\end{figure}

Despite many efforts, existing STISR methods typically treat the character regions and non-character background equally in their forward process while neglecting the adverse impact of the complex background. Intuitively, the non-character background is usually uninformative for the downstream recognition task, so it is unnecessary to reconstruct texture details of the background. Moreover, the complex background can bring disturbance to the reconstruction process. On one hand, the background may be mistakenly treated as characters, thus producing incorrect reconstruction (see Fig. \ref{fig1} (a)). On the other hand, the background may prevent the model from accurately locating characters, thus leading to poor reconstruction results (see Fig. \ref{fig1} (b)). Therefore, existing STISR methods usually suffer from performance degradation due to the complex background, thus limiting practical applications.

To address these issues, we propose LEMMA, a Location Enhanced Multi-ModAl network, to treat character regions and non-character background distinctly. Specifically, we introduce a novel Location Enhancement Module (LEM) to exploit the character location information in the attention map generated by the text recognizer. However, adopting the attention map for character region localization is nontrivial, because the low quality caused by the attention drift \cite{cheng2017focusing} can produce wrong guidance. To this end, we propose the Compression and Expansion strategy to process the raw attention map and thus mitigate the attention drift. We then further select character region features using the feature selection technique, which mitigates background disturbances while reducing the computational complexity of the attention mechanism. In addition, we propose the Multi-modal Alignment Module (MAM) to perform visual-semantic bidirectional alignment, which facilitates better alignment between different modalities in a progressive manner. Finally, we introduce Adaptive Fusion Module (AdaFM) to adapt different guidance to different super-resolution blocks. By modeling character location to distinguish character regions from the background, our method can well handle scene text images with complex background.

Our contributions can be summarized as follows:
\begin{itemize}
    \item We introduce a novel approach with explicit character location modeling to cope with the challenge from the complex background. 
    \item We propose a visual-semantic bidirectional alignment and adaptive fusion strategy to generate and utilize high-level text-specific guidance.
    \item  Experiments on TextZoom and four STR benchmarks show that our method achieves consistently state-of-the-art recognition accuracy.
\end{itemize}

\section{Related Work}
\subsection{Scene Text Image Super-resolution}
Different from single image super-resolution (SISR), whose goal is to improve image quality and obtain favorable visual effects. The main objective of STISR is to obtain easy-to-distinguish images to boost downstream recognition task. Early methods \cite{YongqiangMou2020PlugNetDA,dong2015boosting,wang2019textsr} used CNN architectures to perform STISR tasks. In precisely, TextSR \cite{wang2019textsr} uses adversarial training to enable the model to focus more on textual content. Based on the idea of multi-task learning, Plugnet \cite{YongqiangMou2020PlugNetDA} can obtain a unified feature for super-resolution and recognition. Recently, the TextZoom dataset \cite{WenjiaWang2020SceneTI} was proposed to tackle real-world STISR tasks. And they also proposed TSRN to exploit the sequential nature of scene text images. Benefiting from the global receptive field of attention mechanism, TBSRN \cite{JingyeChen2021SceneTT} utilizes the content-aware loss and the position-aware loss to improve reconstruction results. PCAN \cite{CairongZhao2021SceneTI} improves performance by carefully designing SR blocks. TPGSR \cite{JianqiMa2021TextPG} takes a further step to exploit text prior. TG \cite{chen2022text} shows that fine-grained clues can help yield more distinguishable images. TATT \cite{ma2022text} uses the attention mechanism to work with irregular text images. C3-STISR \cite{MinyiZhao2022C3STISRST} uses clues from three perspectives to introduce better guidance. However, existing methods still treat character regions and background equally in the model design. Although some methods (e.g. TBSRN and TG) enable the model to focus on text by designing related loss, this \textit{implict} manner is hard to observe whether the STISR model really focuses more on text, and incorrect supervision can be generated due to distractions such as attention drift. By contrast, we are the first to consider an \textit{explicit} text focus to handle the challenges posed by complex background.

\subsection{Scene Text Recognition} 

Scene text recognition is to recognize character sequences from scene text images. And it is closely related to scene text image super-resolution. Early text recognition methods used a bottom-up approach, but often suffered from low-resolution, small character, rotated, and illuminated scene text images. Recently, attention-based methods have gained interest because of their promising performance on irregular text. Specifically, SAR \cite{li2019show} uses the 2D attention mechanism to recognize irregular text. RobustScanner \cite{yue2020robustscanner} mitigates attention drift by enhancing positional cues. SRN \cite{yu2020towards} and ABINet \cite{fang2021read} facilitate text recognition by using a language model. MATRN \cite{na2022multi} benefits from visual-semantic multi-modality. MGP-STR \cite{wang2022multi} boosts performance by multi-granularity prediction. Despite these advances, current scene text recognition methods still face challenges on low-resolution scene text images, and adapting models to low-resolution images through data augmentation has been shown to be limited \cite{chen2022text,JingyeChen2021SceneTT}. Therefore, it is necessary to use the STISR method to obtain easily recognizable text images.

\begin{figure*}[t]
\centering
\includegraphics[width=0.85\textwidth]{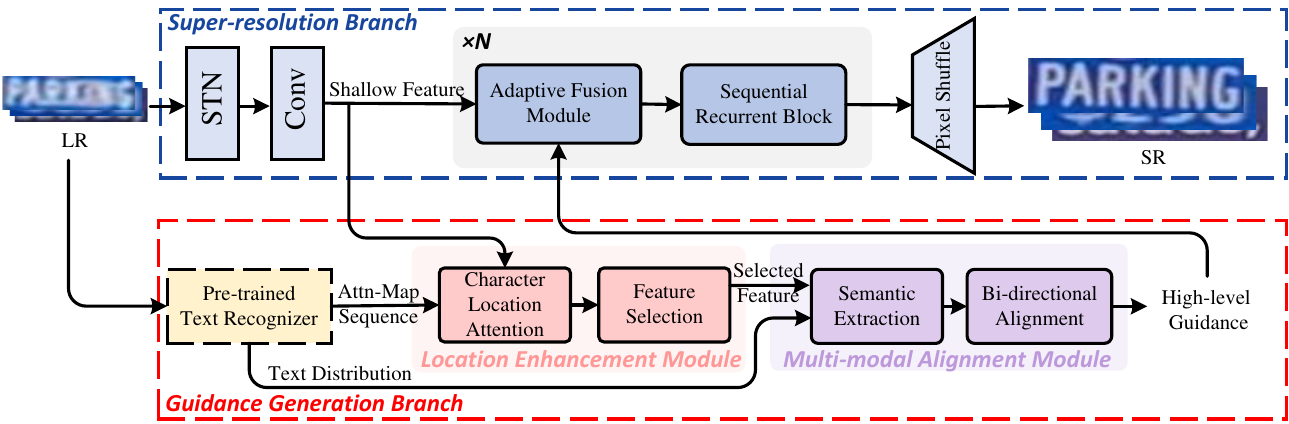}
\caption{Our proposed method consists of two branches, the guidance generation branch uses the text prior to generate high-level guidance which is used in the super-resolution branch to facilitate the reconstruction of high-resolution images.
}
\label{fig2}
\end{figure*}

\section{Methodology}
\label{sec3}
\subsection{Overview}
Given low-resolution image $X_{LR} \in \mathbb{R}^{H\times W\times 3}$, where $H$ and $W$ are the height and width respectively, the goal for STISR is to generate high-resolution text image $X_{SR} \in \mathbb{R}^{fH\times fW\times 3}$, where $f$ is the scale factor.
In the proposed pipeline, $X_{LR}$ is first corrected by Spatial Transformer Network (STN) \cite{MaxJaderberg2015SpatialTN} to tackle the misalignment problem. And then shallow feature $X_I$ is obtained using a shallow CNN. Then, $X_I$ will go through the guidance generation branch and the super-resolution branch. As for the guidance generation branch, the Location Enhancement Module (LEM) takes attention map sequence and $X_I$ as input to generate the selected feature $X_S$. The Multi-modal Alignment Module (MAM) then performs visual-semantic bidirectional alignment using text distribution and $X_S$. As for the super-resolution branch, $X_I$ will go through $N$ stacked blocks, each of which contains an Adaptive Fusion Module (AdaFM) and a Sequential-Recurrent Block (SRB) \cite{WenjiaWang2020SceneTI}. At last, PixelShuffle is performed to increase the spatial scale to generate $X_{SR}$. Fig. \ref{fig2} illustrates the architecture.

\subsection{Location Enhancement Module}
\label{sec:LEM}
\begin{figure}[t]
\centering
\includegraphics[width=0.49\textwidth]{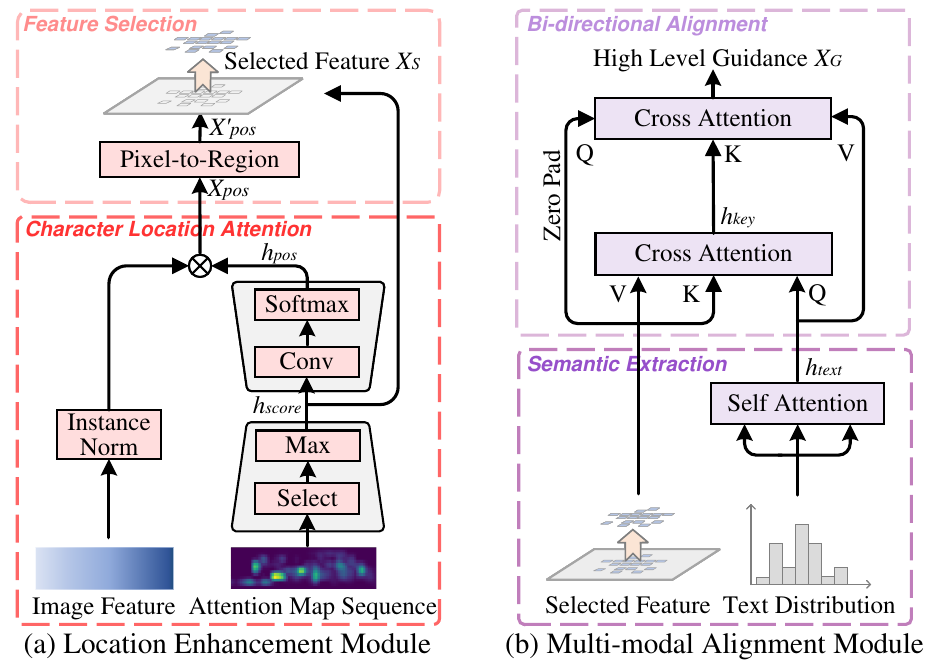}
\caption{The detailed architecture of (a) Location Enhancement Module and (b) Multi-modal Alignment Module. Positional encoding is not represented for brevity.
}
\label{fig3}
\end{figure}

To improve the existing methods' equal treatment between character regions and background, we propose the Location Enhancement Module (LEM) to model the location of each character explicitly (Fig. \ref{fig3} (a)).

\subsubsection{Character Location Attention}
We denote the attention map sequence generated by the pre-trained attention-based text recognizer as $h_{attn} \in \mathbb{R}^{T \times H\times W}$, where $T$ is the max sequence length. Since the character length of different images varies and the raw attention map may introduce misguidance due to attention drift, we thus propose the Compression and Expansion strategy to tackle these problems. 

\noindent
\textbf{Compression Strategy:} Let $L$ be the valid length of one character sequence which can be obtained from the pre-trained recognizer, $h^{j}_{attn}$ denotes the attention map corresponding to $j$-th character. Since different images vary in text lengths, we therefore remove additional paddings by choosing the first $L$ valid attention maps to get $\{h^{j}_{attn}\}_{j=1}^{L}$, and then concatenate them followed by a max operator to reduce the channel dimension to 1. The result is denoted as $h_{score}$:

\begin{equation}
    h_{score} = {\rm{Max}}(
    {\rm Concat}(h^{1}_{attn},\cdots,h^{L}_{attn})),
\end{equation}

\noindent
\textbf{Expansion Strategy:} We then use $C$ convolution kernels to perform up-dimension on $h_{score}$ followed by the Softmax function to get the result $h_{pos}$: 
\begin{equation}
    h_{pos} = {\rm{Softmax}}({\rm{Conv}}(h_{score})),
\end{equation}


\noindent
\textbf{Instance Normalization:} 
According to previous studies \cite{huang2017arbitrary,karras2019style,luo2022siman}, the mean and variance contain the style of an image. To facilitate the subsequent alignment between image and text, we perform Instance Normalization (IN) on $X_{I}$ to remove the varying styles so that focus more on the text content. The normalized features are then multiplied with $h_{pos}$ to obtain the location enhanced feature $X_{pos}$:

\begin{equation}
    X_{pos} = {\rm IN}(X_I)\otimes h_{pos},
\end{equation}

\noindent
where $\otimes$ denotes the Hadamard Product.

\subsubsection{Feature Selection}
An intuition for visual-semantic alignment is that only character regions need to be aligned, while background does not necessarily perform expensive cross attention. Inspired by \cite{tang2022few}, we use the feature selection technique. Unlike \cite{tang2022few} which uses manual annotation as well as a separate scoring network, in this work, we perform feature selection using the easily available attention map. Specifically, since $h_{score}$ contains the pixel-level character confidence, we choose the top K large scores in $h_{socre}$ to get the foreground coordinate set $\mathcal{F}$:

\begin{equation}
    {\mathcal{F}} := \{(m,n) : h_{score}(m,n) \in {\rm TopK}(h_{score})\},
\end{equation}

We then use the coordinates set $\mathcal{F}$ as an index to gather foreground character features from $X_{pos}$. To avoid neighborhood information loss, we apply the pixel-to-region strategy before the gathering to make each indexed pixel representable to its neighbors by weighted summation in local regions:

\begin{equation}
X'_{pos}=\sum_{({\rm \Delta} m,{\rm \Delta} n)\in {\mathcal{N}}}w({\rm \Delta} m, {\rm \Delta} n)X_{pos}(m+{\rm \Delta} m, n+{\rm \Delta} n),
\end{equation}

\noindent
where $\mathcal{N}$ denotes the neighborhood displacements and $w(\cdot,\cdot)$ denotes the weights at each displacement. In the practical implementation, we use the eight-neighborhood region as $\mathcal{N}$. Note that it can be easily implemented by convolution.

Finally, we employ the coordinates in $\mathcal{F}$ to index $X'_{pos}$ and get the selected feature $X_S = \{X'_{pos}(i,j)\}_{(i,j)\in {\mathcal{F}}}$. It is worth noting that the proposed feature selection scheme not only enables more focus on the foreground but also reduces the computational complexity from the attention mechanism, see supplementary material for analysis.

\subsection{Multi-modal Alignment Module}
Existing approaches either use a single modal or use unidirectional multi-modal alignment to generate high-level guidance. We introduce the Multi-modal Alignment Module (MAM) to take a further step toward bidirectional visual-semantic alignment (Fig. \ref{fig3} (b)).

\subsubsection{Semantic Extraction}
We first perform Semantic Extraction on the text distribution obtained by the text recognizer through linear projection and self-attention block. It can generate semantically rich feature $h_{text}$ which will be used to align with the visual modal.

\subsubsection{Bidirectional Alignment}
We propose a bidirectional strategy to facilitate cross-modal alignment. For the image-to-text alignment, we use the $h_{text}$ as query, $X_S$ as key and value to allow each character to find its corresponding image region:

\begin{equation}
\begin{aligned}
&h'_{n} = {\rm LN}({\rm MHA}(h_{n-1},X_S,X_S) + h_{n-1}),\\
&h_{n} =  {\rm LN}({\rm MLP}(h'_{n}) + h'_{n}),
\end{aligned}
\end{equation}

\noindent
where $n$ denotes the $n$-th attention block. $h_{n-1}$ is $h_{text}$ if $n=1$ otherwise the output of previous block.

We denote the result of the first level alignment as $h_{key}$. Note that $h_{key}$ has the same size as $h_{text}$ but contains information of $X_S$. So, it can be easily aligned with $X_S$ in the second level alignment. For text-to-image alignment, we use $h_{key}$ to bridge visual and semantic modals. Specifically, we use $X_{S}$ as query, $h_{key}$ as key, and $h_{text}$ as value. Each element of $X_{S}$ can find which text feature it should attend by using $h_{key}$. Since the output of attention shares the same size as query, for the subsequent fusion, we use zero to pad $X_S$ before the second level alignment. The output high-level guidance $X_{G}$ is then used to guide super-resolution. Moreover, we refer to the positional encoding in \cite{na2022multi} to make full use of the attention map for better alignment.

It is worth noting that the previous unidirectional alignment only uses the second-level cross attention with $h_{text}$ as key and value \cite{ma2022text}. However, without the progressive transition from the first level, $X_S$ (query) and $h_{text}$ (key) are difficult to align well. Related experiment can be seen in Section \ref{sec:bi-direction}.

\subsection{Adaptive Fusion Module}

Current guidance-based STISR approaches \cite{JianqiMa2021TextPG,MinyiZhao2022C3STISRST,ma2022text} use the same guidance in different SR blocks. For this reason, we propose the Adaptive Fusion Module (AdaFM) to adaptively incorporate high-level guidance to different blocks. Given image feature $X_{n}$ which is $X_I$ or the output of the previous block, and high-level guidance $X_G$, we first concatenate them along channel dimension followed by three parallel $1 \times 1$ convolution to project $X_{n}$ into three different feature spaces and denote them as $X_{n}^1$, $X_{n}^2$, and $X_{n}^3$ respectively. We then perform the channel attention mechanism on $X_{n}^1$ and multiply the resulting score with $X_{n}^2$ to generate the channel attention feature, which will be added to $X_{n}^3$ to get the final result.
Notably, unlike the previous channel attention \cite{hu2018squeeze}, we use global deep-wise convolution \cite{chollet2017xception} rather than global average pooling to better exploit the property of the spatial distribution of character regions in scene text images. The process can be formalized as follows:

\begin{equation}
    X_{n+1} = X_{n}^3+X_{n}^2 \otimes {\rm Sigmoid}({\rm MLP}({\rm GDWConv}(X_{n}^1))),
\end{equation}

\noindent
where GDWConv denotes the Global Deep-Wise Convolution. The above procedure only uses $1\times1$ convolution and therefore results in only a minimal increase in the parameter complexity.

\subsection{Training Objective}

In this work, we use three loss functions, namely pixel loss, recognition loss, and fine-tuning loss, to train our model.

For the pixel loss, we use the $L_2$ loss to perform pixel-level supervision: 

\begin{equation}
L_{pix} = || X_{SR}-X_{HR}||_2,
\end{equation}

For the recognition loss, we use the text-focus loss \cite{JingyeChen2021SceneTT} to supervise the learning of language knowledge:

\begin{equation}
    L_{txt} = \lambda_1||\textbf{A}_{HR}-\textbf{A}_{SR}||_1+\lambda_2 \text{WCE}(p_{SR}, y_{label}),
\end{equation}

\noindent
where $\textbf{A}$ and $p$ are the attention map and probability distribution predicted by a fixed transformer-based recognizer, respectively. WCE denotes weighted cross-entropy. $\lambda_1$ and $\lambda_2$ are hyperparameters.

Since we use a pre-trained text recognizer for guidance generation, as \cite{ma2022text,MinyiZhao2022C3STISRST} demonstrated, fine-tuning is better than fixed parameters. We therefore use the fin-tuning loss to adapt the text recognizer to low-resolution inputs:

\begin{equation}
    L_{ft} = \text{CE}(p_{PRE},y_{label}),
\end{equation}

\noindent
where CE denotes cross-entropy loss, $p_{PRE}$ denotes probability distribution predicted by the pre-trained text recognizer.

The total loss is the weighted sum of these losses above.
\begin{equation}
\label{loss}
L = L_{pix} + \alpha_1 L_{txt} + \alpha_2 L_{ft},    
\end{equation}

\noindent
where $\alpha_1$ and $\alpha_2$ are hyperparameters.

\section{Experiment}

\subsection{Datasets}

\subsubsection{Scene Text Image Super-resolution Dataset} 
TextZoom \cite{WenjiaWang2020SceneTI} is widely used in STISR works. This dataset is derived from two single image super-resolution datasets, RealSR \cite{JianruiCai2019TowardRS} and SR-RAW \cite{XuanerZhang2019ZoomTL}. The images are captured by digital cameras in real-world scenes. In total, TextZoom contains 17367 LR-HR pairs for training and 4373 pairs for testing. Depending on the focal length of the digital camera, the test set is divided into three subsets, with 1619 pairs for the simple subset, 1411 pairs for the medium subset, and 1343 pairs for the hard subset. The size of LR images is adjusted to 16 × 64, and the size of the HR images is 32 × 128.

\subsubsection{Scene Text Recognition Datasets}
To verify the robustness of our method in the face of different styles of text images, we adopt four scene text recognition benchmarks, namely ICDAR2015 \cite{DimosthenisKaratzas2015ICDAR2C}, CUTE80 \cite{PalaiahnakoteShivakumara2014ARA}, SVT \cite{KaiWang2011EndtoendST} and SVTP \cite{TrungQuyPhan2013RecognizingTW} to evaluate our model. Since these datasets do not contain LR-HR pairs and most of the images are of high quality, we therefore first perform preprocess to get LR images. Specifically, we select images with less than $16 \times 64$ pixels and then perform manual degradation on these selected images. One can find more information about these datasets' descriptions and degradation details in the supplementary material.


\subsection{Evaluation Metrics} For the text recognition accuracy evaluation, consistent with previous work \cite{JianqiMa2021TextPG,WenjiaWang2020SceneTI,MinyiZhao2022C3STISRST,ma2022text}, we use three text recognizers namely CRNN \cite{shi2016end}, MORAN \cite{luo2019moran}, and ASTER \cite{shi2018aster} for evaluation in order to guarantee the generalization. For image fidelity evaluation, we used Peak Signal-to-Noise Ratio (PSNR) and Structural Similarity Index Measure (SSIM) to assess the quality of the generated SR images. However, as found in previous work \cite{chen2022text,JingyeChen2021SceneTT,MinyiZhao2022C3STISRST,CairongZhao2021SceneTI}, there is an \textit{inherent trade-off} between text recognition accuracy and image quality, and we will discuss this observation in Section \ref{sec-trade-off}.

\subsection{Implementation Details}
We use ABINet \cite{fang2021read} as the attention-based text recognizer because of the accessibility of code and pre-trained model. We train our model with batch size 64 for 500 epochs using Adam \cite{kingma2014adam} for optimization. The learning rate is set to 1e-3 for the super-resolution and 1e-4 for fine-tuning ABINet, both are decayed with a factor of 0.5 after 400 epochs. We refer to the hyperparameters on $L_{txt}$ given in \cite{JingyeChen2021SceneTT}, namely $\lambda_1 = 10$, $\lambda_2 = 0.0005$. For the other hyperparameters, we use $\alpha_1 = 0.5$, $\alpha_2 = 0.01$, see supplementary material for details.

\subsection{Ablation Study}

In this section, we conduct an ablation study to demonstrate the effectiveness of each component in LEMMA. CRNN \cite{shi2016end} is chosen as the text recognizer for uniformity.

\subsubsection{Different Choices on Location Enhancement} In the Location Enhancement Module (LEM), we use character location attention and feature selection to enable explicit character location modeling. Table \ref{LEM} shows the effect of each part. The presence of character location attention resulted in a 3.9\% improvement compared to no location enhancement. When using feature selection, recognition accuracy does not decrease due to the neglect of the background feature, instead it further increases by 2.2\%. This suggests that there is redundant information in the feature map and removing this redundancy will improve performance.

\begin{table}[]
\centering
\resizebox{\columnwidth}{!}{%
\begin{tabular}{cc|cccc}
\hline
\multicolumn{2}{c|}{Location Enhance}                   & \multicolumn{4}{c}{Recognition Accuracy} \\ \hline
cha-attn                   & feat-select                & Easy   & Medium   & \multicolumn{1}{c|}{Hard} & avgAcc  \\ \hline
-             & -        & 60.1\% & 50.5\% &  \multicolumn{1}{c|}{38.0\%}&50.2\% \\
-             & \ding{51}& 62.1\% & 51.2\% &    \multicolumn{1}{c|}{38.2\%} & 51.3\% \\
\ding{51} & -            & 65.0\% & 55.3\%    &  \multicolumn{1}{c|}{39.8\%}& 54.1\% \\
\ding{51} & \ding{51}    & \textbf{67.1\%} & \textbf{58.8\%} &\multicolumn{1}{c|}{\textbf{40.6\%}} & \textbf{56.3\%} \\ \hline
\end{tabular}%
}
\caption{Ablation study on LEM. \textit{cha-attn} and \textit{feat-select} denote the use of character location attention and feature selection, respectively.}
\label{LEM}
\end{table}

\begin{table}[]
\centering
\resizebox{\columnwidth}{!}{%
\begin{tabular}{c|cccc}
\hline
\multirow{2}{*}{Aligment Strategy} & \multicolumn{4}{c}{Recognition Accuracy}
\\ \cline{2-5} 
                    & Easy & Medium & \multicolumn{1}{c|}{Hard} & avgAcc \\ \hline
NoA                  &  63.5\%  & 53.4\% & \multicolumn{1}{c|}{37.4\%} & 52.2\%      \\
UDA                  & 64.2\% & 56.7\%   & \multicolumn{1}{c|}{\textbf{41.2\%}} & 54.5\%  \\
BDA                  & \textbf{67.1\%}  & \textbf{58.8\%}  & \multicolumn{1}{c|}{40.6\%}&\textbf{56.3\%}     \\ \hline
\end{tabular}%
}
\caption{Ablation study on different alignment strategies. NoA indicates No Alignment with only text modal, UDA indicates UniDirectional Alignment, and BDA indicates BiDirectional Alignment.}
\label{MAM}
\end{table}

\begin{table*}[t]
\resizebox{\textwidth}{!}{%
\begin{tabular}{c|cccc|cccc|cccc}
\hline
\multirow{2}{*}{Method} & \multicolumn{4}{c|}{ASTER \cite{shi2018aster}}         & \multicolumn{4}{c|}{MORAN \cite{luo2019moran}}         & \multicolumn{4}{c}{CRNN \cite{shi2016end}}                    \\ \cline{2-13} 
                        & Easy   & Medium & Hard   & Average & Easy   & Medium & Hard   & Average & Easy            & Medium & Hard   & Average \\ \hline
BICUBIC                 & 67.4\% & 42.4\% & 31.2\% & 48.2\%  & 60.6\% & 37.9\% & 30.8\% & 44.1\%  & 36.4\%          & 21.1\% & 21.1\% & 26.8\%  \\
HR                      & 94.2\% & 87.7\% & 76.2\% & 86.6\%  & 91.2\% & 85.3\% & 74.2\% & 84.1\%  & 76.4\%          & 75.1\% & 64.6\% & 72.4\%  \\ \hline
SRCNN \cite{dong2015image} & 70.6\% & 44.0\% & 31.5\% & 50.0\%  & 63.9\% & 40.0\% & 29.4\% & 45.6\%  & 41.1\%          & 22.3\% & 22.0\% & 29.2\%  \\
SRResNet \cite{ledig2017photo}  & 69.4\% & 50.5\% & 35.7\% & 53.0\%  & 66.0\% & 47.1\% & 33.4\% & 49.9\%  & 45.2\%          & 32.6\% & 25.5\% & 35.1\%  \\
RCAN \cite{zhang2018image}  & 67.3\% & 46.6\% & 35.1\% & 50.7\%  & 63.1\% & 42.9\% & 33.6\% & 47.5\%  & 46.8\%          & 27.9\% & 26.5\% & 34.5\%  \\
SAN \cite{dai2019second} & 68.1\% & 48.7\% & 36.2\% & 50.7\%  & 65.6\% & 44.4\% & 35.2\% & 49.4\%  & 50.1\%          & 31.2\% & 28.1\% & 37.2\%  \\
TSRN \cite{WenjiaWang2020SceneTI} & 75.1\% & 56.3\% & 40.1\% & 58.3\%  & 70.1\% & 55.3\% & 37.9\% & 55.4\%  & 52.5\%          & 38.2\% & 31.4\% & 41.4\%  \\
TBSRN \cite{JingyeChen2021SceneTT} & 75.7\% & 59.9\% & 41.6\% & 60.1\%  & 74.1\% & 57.0\% & 40.8\% & 58.4\%  & 59.6\%          & 47.1\% & 35.3\% & 48.1\%  \\
PCAN \cite{CairongZhao2021SceneTI}    & 77.5\% & 60.7\% & 43.1\% & 61.5\%  & 73.7\% & 57.6\% & 41.0\% & 58.5\%  & 59.6\%          & 45.4\% & 34.8\% & 47.4\%  \\
TG \cite{chen2022text}    & 77.9\% & 60.2\% & 42.4\% & 61.3\%  & 75.8\% & 57.8\% & 41.4\% & 59.4\%  & 61.2\%          & 47.6\% & 35.5\% & 48.9\%  \\
TATT \cite{ma2022text}   & 78.9\% & 63.4\% & 45.4\% & 63.6\%  & 72.5\% & 60.2\% & 43.1\% & 59.5\%  & 62.6\%          & 53.4\% & 39.8\% & 52.6\%  \\
C3-STISR \cite{MinyiZhao2022C3STISRST}   & 79.1\% & 63.3\% & 46.8\% & 64.1\%  & 74.2\% & 61.0\% & 43.2\% & 59.5\%  & 65.2\% & 53.6\% & 39.8\% & 53.7\%  \\ \hline
LEMMA (Ours) &
  \textbf{81.1\%} &
  \textbf{66.3\%} &
  \textbf{47.4\%} &
  \textbf{66.0\%} &
  \textbf{77.7\%} &
  \textbf{64.4\%} &
  \textbf{44.6\%} &
  \textbf{63.2\%} &
  \textbf{67.1\%} &
  \textbf{58.8\%} &
  \textbf{40.6\%} &
  \textbf{56.3\%} \\ \hline
\end{tabular}%
}
\caption{Comparison of the downstream text recognition accuracy on the TextZoom dataset. The best result is bolded.}
\label{comparision}
\end{table*}

\subsubsection{Effectiveness of Bidirectional Alignment Strategy} 
\label{sec:bi-direction}
To demonstrate the performance improvement does come from the bidirectional strategy rather than additional parameters, we increase the parameter number in the unidirectional alignment strategy by stacking more layers to make the complexity consistent. The results in Table \ref{MAM} show that the bidirectional alignment improves the average accuracy by 1.8\% compared to its unidirectional counterpart, demonstrating the effectiveness of the bidirectional alignment strategy.

\subsubsection{Effectiveness of Different Modules}
We study the effectiveness of different modules in LEMMA. Table \ref{ablation:combine} shows the result. It can be seen that the simultaneous presence of the three modules achieves the best result.

\begin{table}[]
\centering
\resizebox{\columnwidth}{!}{
\begin{tabular}{ccc|ccc|c}
\hline
\multirow{2}{*}{LEM} & \multirow{2}{*}{MAM} & \multirow{2}{*}{AdaFM} & \multicolumn{4}{c}{Recognition Accuracy} \\ \cline{4-7} 
                     &                      &                        & Easy    & Medium    & Hard    & avgAcc   \\ \hline
-                    & -              & -            & 61.1\%   & 50.0\%     & 36.4\%    & 50.0\%       \\
\ding{52}            & -              &-    &  65.6\%   & 55.6\%  & 40.4\%      & 54.6\%   \\
\ding{52}           & \ding{52}      & -    & 66.6\%     & 56.3\%  & \textbf{41.3\%}     & 55.5\%      \\
\ding{52}      & \ding{52}      & \ding{52} &\textbf{67.1\%} & \textbf{58.8\%}  & 40.6\%       & \textbf{56.3\%}    \\ \hline
\end{tabular}%
}
\caption{Combination of different components in LEMMA. Concatenate is used to perform fusion when AdaFM is removed.}
\label{ablation:combine}
\end{table}

\subsection{Comparison to State-of-the-Arts}

We first compare the proposed method with others on the TextZoom dataset. Consistent with the previous methods, we evaluate the generalization using three text recognizers, namely ASTER \cite{shi2018aster}, MORAN \cite{luo2019moran}, and CRNN \cite{shi2016end}. After that, we evaluate the model robustness on more challenging STR datasets.

\subsubsection{Results on TextZoom}
\begin{figure*}[]
\centering
\includegraphics[width=\textwidth]{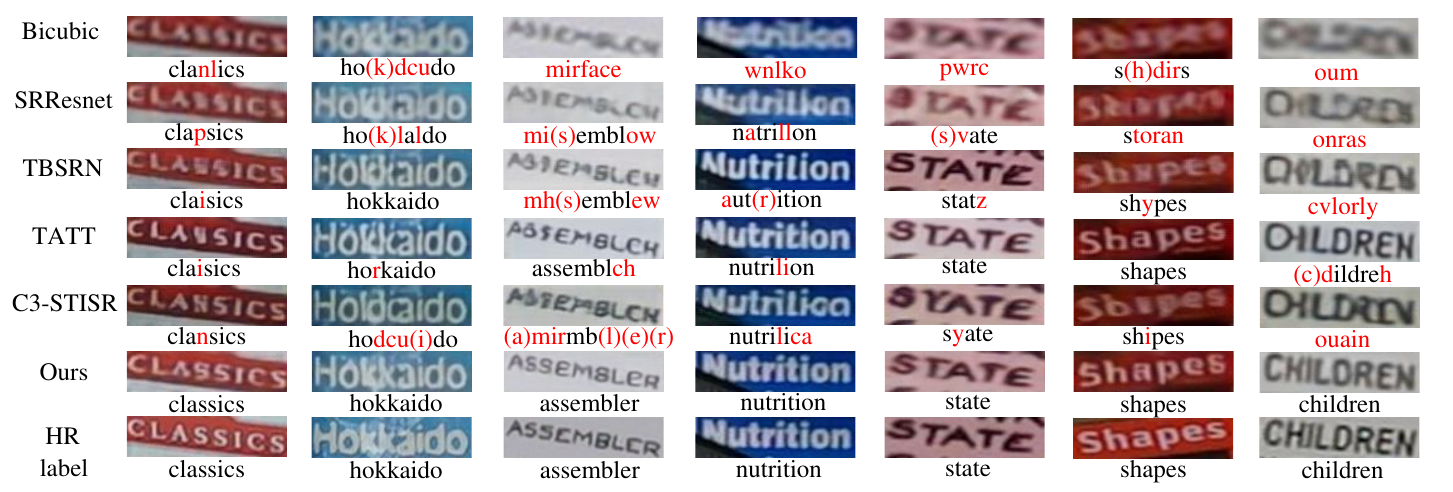}
\caption{Qualitative comparison with other methods. Zoom in for better visualization.}
\label{compare-fig}
\end{figure*}

Table \ref{comparision} shows the comparison results in terms of text recognition accuracy. It can be seen that our proposed LEMMA consistently outperforms the previous method on all three text recognizers. For example, using ASTER as the recognizer, compared to the previous SoTA method \cite{MinyiZhao2022C3STISRST}, our method achieves 2.0\%, 3.0\%, and 0.6\% improvement on easy, medium, and hard, respectively,  which ultimately leads to an average improvement of 1.9\%. The good results demonstrate the advantages of the proposed model. We also present the qualitative comparison results in Fig. \ref{compare-fig}.

\begin{table}[]
\resizebox{\columnwidth}{!}{%
\begin{tabular}{c|cccc}
\hline
\multirow{2}{*}{Method} & \multicolumn{4}{c}{STR Datasets} \\ \cline{2-5} 
                        & IC15  & CUTE80 & SVT  & SVTP \\ \hline
Bicubic                 & 9.5\%   & 35.8\%    & 3.3\%   & 10.2\%  \\
SRResnet                & 13.0\%  & 48.3\%   & 9.3\%   & 12.1\%  \\
TBSRN                   & 20.7\%  & 75.0\%    & 12.2\%  & 17.4\%  \\
TATT                    & 28.6\%  & 74.0\%  & 14.0\%  & 25.9\% \\
C3-STISR                & 22.7\%  & 71.5\%  & 10.2\%  & 17.7\%  \\ \hline
LEMMA (Ours)             & \textbf{32.5\%} & \textbf{76.0\%}    & \textbf{21.3\%}  & \textbf{28.4\%}  \\ \hline
\end{tabular}%
}
\caption{Comparison results on scene text recognition benchmarks.}
\centering
\label{STR-table}
\end{table}

\subsubsection{Towards Robust STISR}
In this section, we explore the model robustness in more challenging samples. All models are trained on the TextZoom dataset, after which we freeze parameters and evaluate them on four text recognition benchmarks namely ICDAR2015 \cite{DimosthenisKaratzas2015ICDAR2C}, CUTE80 \cite{PalaiahnakoteShivakumara2014ARA}, SVT \cite{KaiWang2011EndtoendST}, and SVTP \cite{TrungQuyPhan2013RecognizingTW}. 

Since most of the samples in these datasets contain real-world complex background, the result can thus give an indication of the model's ability to cope with complex background. As shown in Table \ref{STR-table}, our proposed method allows the model to focus more on character regions and leads to better results. 

\begin{figure}[]
\centering
\includegraphics[width=0.46\textwidth]{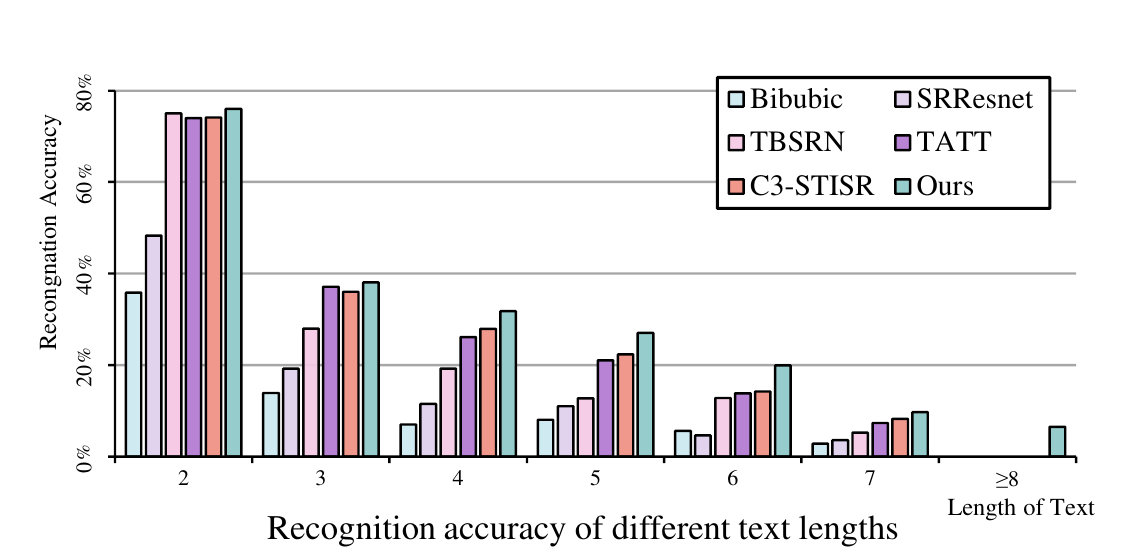}
\caption{Recognition accuracy with different text lengths on four STR benchmarks.}
\label{text-len}
\end{figure}

Moreover, we also use all of the above four STR benchmarks to compare accuracy with different text lengths. It can be seen in Fig. \ref{text-len} that our method outperforms others across all text lengths, and shows strong robustness in dealing with text of extreme length. The favorable performance of our approach on long text instances stems from the fact that explicit location enhancement allows the model to reinforce character region features and thus alleviate the long-range forgetting problem associated with long text.

\begin{table}[]
\setlength{\tabcolsep}{5mm}{
\resizebox{\columnwidth}{!}{%
\begin{tabular}{cc|cc|c}
\hline
\multirow{2}{*}{LEM} & \multirow{2}{*}{RecLoss} & \multicolumn{2}{c|}{Fidelity} & Accuracy \\ \cline{3-5} 
                     &                           & PSNR      & SSIM       & avgAcc   \\ \hline
-                    & -                         & \textbf{21.2}  & 0.7749    & 44.6\%    \\
\ding{52}                  & -         & 20.8         & 0.7708         & 49.5\%    \\
-                    & \ding{52}        & 21.1         & 0.7649        & 50.2\%    \\
\ding{52}                  & \ding{52}  & 20.9         & \textbf{0.7792}         & \textbf{56.3\%}    \\ \hline
\end{tabular}%
}
}
\caption{Trade-off between Fidelity and Accuracy. \textit{RecLoss} denotes the recognition loss.}
\centering
\label{trade-off}
\end{table}

\begin{table}[]
\centering
\resizebox{\columnwidth}{!}{%
\begin{tabular}{cc|ccc|c}
\hline
\multirow{2}{*}{ExpsConv} & \multirow{2}{*}{Pix2Reg} & \multicolumn{4}{c}{Accuracy}   \\ \cline{3-6} 
                    &                          & Easy & Medium & Hard  & avgAcc \\ \hline
-                   & -                        & 62.8\% & 56.5\% & \textbf{41.3\%} & 54.2\%    \\
-                  & \ding{52}                  & 65.5\% & 55.8\% & 40.5\% &  54.7\%   \\
\ding{52}                   & -            & 63.6\% & 50.7\%  & 38.1\% & 51.6\%      \\
\ding{52}            & \ding{52}   &\textbf{67.1\%}  & \textbf{58.8\%} & 40.6\%  & \textbf{56.3\%}   \\ 
\hline
\end{tabular}%
}
\caption{Effectiveness of strategies to mitigate attention drift. \textit{ExpsConv} represents the use of Expansion strategy, \textit{Pix2Reg} denotes the utilization of neighborhood information in feature selection.}
\label{dis:attention-drift}
\end{table}

\subsection{Discussion}
\subsubsection{Trade-off between Fidelity and Accuracy}
\label{sec-trade-off}
As presented in previous studies \cite{JingyeChen2021SceneTT,chen2022text,MinyiZhao2022C3STISRST}, we also find the inherent trade-off between fidelity and accuracy. Table \ref{trade-off} shows this observation. It can be seen that the highest PSNR is achieved with the settings of no LEM and recognition loss. The presence of LEM leads to a 0.4 dB reduction in PSNR, but it gives a 4.9\% improvement in accuracy. The same phenomenon also happens with recognition loss. It can be explained by the fact that the proposed LEM mainly focuses on the foreground character regions and thus does not reconstruct enough of the background that occupies most of one image. Therefore, further recognition accuracy improvements (e.g. generating images with larger contrast) will come at the expense of fidelity.

\subsubsection{Performance Brought from a Strong Recognizer}
In this work, we use the ABINet to generate guidance instead of CRNN as in other works \cite{ma2022text,MinyiZhao2022C3STISRST}. Since ABINet itself can provide more reliable guidance, we conduct experiments to investigate how much the accuracy improvement is attributed to the use of a stronger recognizer. Specifically, we replace ABINet with CRNN to generate text distribution, and the average recognition accuracy is 55.2\% (CRNN for downstream recognition). The adoption of a weak recognizer caused a 1.1\% decrease compared with LEMMA. However, this setting still exceeds SoTA method (53.7\%), demonstrating the validity of explicitly distinguishing between character regions and background.

\subsubsection{Effectiveness of Strategies to Mitigate Attention Drift}
\label{sec:attention-drift}
We distinguish character regions from background using the attention map. As discovered in previous methods \cite{cheng2017focusing,yue2020robustscanner}, attention drift can affect the quality of the attention map and thus produce misguidance. Therefore, solutions are proposed in our framework to mitigate this adverse effect. First, in character location attention, we do not use the raw attention map directly, but process it with the convolution layer which can serve as a shift operator and thus correct the shifted attention map. Second, in feature selection, we use the pixel-to-region strategy so that each pixel can represent its neighbors and thus mitigate the impact of drift. The results in Table \ref{dis:attention-drift} indicate that all these strategies can alleviate the adverse effects from attention drift.

\subsubsection{Failure Case and Limitation}
\begin{figure}[]
\centering
\includegraphics[width=0.45\textwidth]{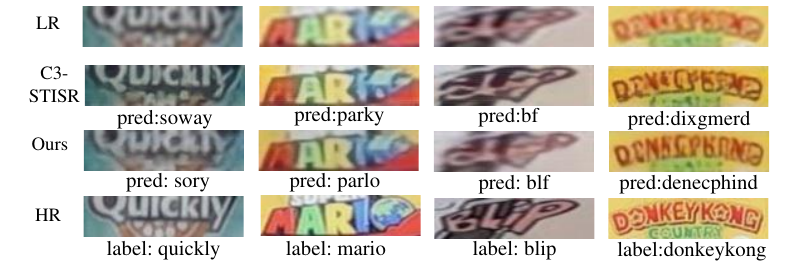}
\caption{Some failure cases of our proposed model.}
\label{faulure-case}
\end{figure}

Fig. \ref{faulure-case} shows some failure cases. Our method, including previous STISR methods, has difficulty handling incomplete characters and artistic characters. Moreover, the quality of attention map which is used as prior guidance can affect the reconstruction results. Despite some strategies are proposed to mitigate this problem, it is far from being completely solved and we leave it as future work.

\section{Conclusion}
In this work, we propose the Location Enhanced Multi-ModAl network (LEMMA) to handle challenges in existing STISR methods by explicit location enhancement with more focus on character regions. The Location Enhancement Module extracts character region features from all pixels through character location attention and feature selection techniques. The Multi-modal Alignment Module employs a bidirectional progressive strategy to facilitate cross-modal alignment. The Adaptive Fusion Module adaptively incorporates the generated high-level guidance into different super-resolution blocks. Results on TextZoom and four challenging STR benchmarks show that our approach consistently improves downstream recognition accuracy, taking a further step toward robust scene text image super-resolution.

\bibliographystyle{named}
\bibliography{ijcai23}

\end{document}